\documentclass{article}

\usepackage{natbib}
\usepackage{PRIMEarxiv}
\usepackage{tikz}
\usetikzlibrary{arrows.meta,positioning,fit,calc}

\usepackage[utf8]{inputenc} %
\usepackage[T1]{fontenc}    %
\usepackage{hyperref}       %
\usepackage{xurl}           %
\usepackage{booktabs}       %
\usepackage{amsfonts}       %
\usepackage{nicefrac}       %
\usepackage{microtype}      %
\usepackage{lipsum}
\usepackage{fancyhdr}       %
\usepackage{graphicx}       %
\graphicspath{{media/}}     %

\usepackage{amsmath}
\usepackage{tabularx}
\usepackage{todonotes}
\usepackage{multirow}
\usepackage{algorithm}
\usepackage{algpseudocode}

\title{From Shots to Rallies to Sets: Hierarchy-Aligned Multi-Task Learning for Next-Shot Prediction in Racket Sports}

\author{
  Chao-Wei Cai$^\dag$, Min-Te Sun$^\dag$, Sheng Kuang Wu$^\ddag$, Hung-Hsuan Chen$^\dag$\\
  $^\dag$Computer Science and Information Engineering, National Central University, Taoyuan, Taiwan \\
  $^\ddag$Department of Kinesiology, National Tsing Hua University, Hsinchu, Taiwan \\
  \texttt{a96521852@gmail.com, msun@csie.ncu.edu.tw, skwu@mx.nthu.edu.tw, hhchen1105@acm.org}
}

\begin{document}
\maketitle

\begin{abstract}
Next-shot prediction in racket sports requires learning from event data with different match structures, recorded attributes, and label vocabularies. Table tennis, badminton, and tennis nevertheless share a hierarchy in which shots form rallies and rallies form higher scoring units. We present MT-HTA, a hierarchy-aligned multi-task Transformer that uses this structure to jointly predict the attributes of the next shot. A hierarchical temporal Transformer captures event progression at each level of the hierarchy, a feature-token path models interactions among semantic fields rather than along the time axis, and task-specific attention retrieves information from hierarchical summaries and player context. The architecture supports three- and four-level hierarchies and sport-dependent task sets. Across four datasets from three sports, the final MT-HTA achieves the highest mean task-averaged F1 among the compared main models. Ablations on table tennis support the combination of hierarchical encoding, task attention, and feature-token interaction. Additional modules alter the balance between classification scores and spatial error rather than improving every metric. These results support a reusable architecture for heterogeneous racket-sport event data. We release the code for reproducibility.\footnote{\url{https://github.com/gndd1221/For-Sports-event-modeling-with-rally-and-score-hierarchical-A-Multi-Task-Hierarchical-Transformer}}
\end{abstract}


\section{Introduction} \label{sec:intro}

Racket sports such as table tennis, badminton, and tennis are fast, continuous decision-making games with explicit rally boundaries, and predicting a player's next shot---its type, placement, hand, strength, and spin---is valuable for tactical analysis, opponent modeling, and training support. They also share a common rule-induced hierarchy: a shot is part of a rally, rallies form games or sets, and the next decision depends on both the short-term exchange within the current rally and the longer-term match state. Professional annotation systems, crowdsourced charting projects, and automated vision pipelines now log every shot of every rally, turning what used to be a handful of hand-coded matches into a continuously growing, spatio-temporal event stream: the four corpora we use already hold more than $240,000$ annotated shots, and each of the three sports contributes records under a different schema, label vocabulary, and structural depth. Next-shot prediction is therefore no longer only a modeling question; it is a question of how to learn over heterogeneous big data without rebuilding the stack for every source.
 
Three gaps stand in the way. First, existing stroke-forecasting models~\citep{Wang2022ShuttleNet, Chang2023DyMF} are built one sport at a time and typically optimize one or two attributes, leaving the open question of whether a single architecture can serve sports with different rules, hierarchy depths, and label spaces. This is the classic \emph{variety} bottleneck of big data analytics: covering a new sport, or the same sport under a different annotation provider, means designing, preprocessing, tuning, and maintaining another model from scratch. Second, the dominant practice of flattening a match into one long sequence---whether recurrent or Transformer-based~\citep{Hochreiter1997LSTM, Vaswani2017}---erases the rally and set boundaries the rules impose, forcing attention across tactically discontinuous shots and blending short-term exchanges with long-term match structure. Third, and most fundamentally, the literature rarely asks what actually generalizes across racket sports: which parts of a hierarchical model carry over unchanged from sport to sport, and which are sport-specific adaptations.
 
We believe that next-shot prediction is best posed as a single, structurally grounded, multi-task problem shared across racket sports. Concretely, we make the following claim.
 
\paragraph{Central claim}
A hierarchy-aligned multi-task architecture can be instantiated with a single formulation and training recipe across racket sports whose rules, hierarchy depth, and label spaces differ, while remaining competitive with generic sequence baselines on each sport.
 
\paragraph{Why this is a machine-learning-on-big-data problem}
Three properties of the data drive the design. \emph{Variety} dominates. Three sports, four corpora, and three independent annotation protocols disagree on at least three things: how deep the match structure is ($D=3$ for table tennis and badminton, $D=4$ for tennis); which shot attributes are annotated (badminton records neither strength nor spin and tennis omits strength); and how many classes each attribute has, the shot-type vocabulary alone ranging from $10$ to $19$. Thus, we lack a representation general enough to ingest all of them. \emph{Volume} varies across corpora: they span $45,714$ to $108,289$ shots, and the expanded table tennis corpus enlarges the original one by roughly $2.3\times$ in matches, players, and shots, letting us evaluate the same model at two data scales. \emph{Velocity} shapes the interface. The model reads the shots already played in the current rally, together with the rallies and the games or sets that have already finished, and emits every attribute in one forward pass. That is the shape an in-match system needs, and a single prediction costs only milliseconds on a laptop CPU. Multi-task learning (one shared encoder amortized over every attribute of every sport) is what makes this affordable.

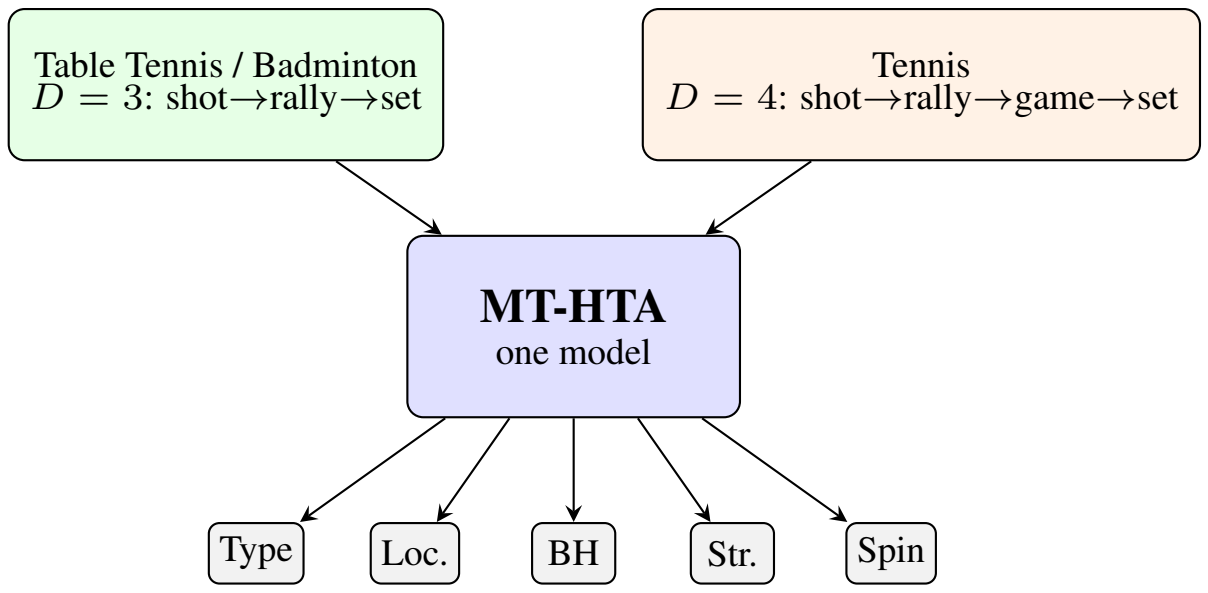
\begin{figure}[t]
\centering
\begin{tikzpicture}[
    sport/.style={draw, rounded corners=3pt, align=center,
                  minimum height=1.0cm, minimum width=2.5cm,
                  font=\scriptsize, inner sep=4pt},
    model/.style={draw, rounded corners=3pt, align=center,
                  minimum height=1.2cm, minimum width=2.2cm,
                  font=\small, fill=blue!12, inner sep=4pt},
    task/.style={draw, rounded corners=2pt, align=center,
                 minimum height=0.4cm, minimum width=0.55cm,
                 font=\scriptsize, fill=gray!10, inner sep=2pt},
    arr/.style={->, >=stealth, thin},
    lvl/.style={font=\scriptsize, fill=yellow!20, draw, rounded corners=1pt,
                inner sep=2pt},
]
\node[sport, fill=green!10] (tt) at (-2.3,0)
    {Table Tennis / Badminton\\[-2pt]\scriptsize $D=3$: shot$\to$rally$\to$set};
\node[sport, fill=orange!10] (tn) at (2.3,0)
    {Tennis\\[-2pt]\scriptsize $D=4$: shot$\to$rally$\to$game$\to$set};
\node[model] (mthta) at (0,-1.6) {\textbf{MT-HTA}\\[-2pt]\scriptsize one model};
\node[task] (t1) at (-2.1,-3.1) {Type};
\node[task] (t2) at (-1.05,-3.1) {Loc.};
\node[task] (t3) at (0,-3.1)    {BH};
\node[task] (t4) at (1.05,-3.1) {Str.};
\node[task] (t5) at (2.1,-3.1)  {Spin};
\draw[arr] (tt) -- (mthta);
\draw[arr] (tn) -- (mthta);
\foreach \t in {t1,t2,t3,t4,t5} { \draw[arr] (mthta) -- (\t); }
\end{tikzpicture}
\caption{MT-HTA applies one formulation and training recipe to sports with different hierarchy depths, jointly predicting each shot's attributes. It matches or exceeds generic sequence baselines across all four datasets (Table~\ref{tab:main}), with gains driven by a shared core rather than by sport-specific components.}
\label{fig:teaser}
\end{figure}

Figure~\ref{fig:teaser} summarizes the resulting arrangement: sports of different hierarchy depth enter one model that predicts every attribute of the next shot jointly.

\paragraph{Contributions}
\begin{enumerate}
  \item A unified hierarchy-aligned multi-task formulation that admits a variable hierarchy depth $D\in\{3,4\}$ and a variable task set without per-sport customization, so supporting a new source is a matter of mapping its schema, not of redesigning the model.
  \item Cross-sport evidence: under one training recipe, the highest mean Macro F1 on all four datasets, with margins over the strongest baselines quantified across three seeds.
  \item A component analysis that, through controlled ablations, isolates the shared core (hierarchy, task attention, and feature-token interaction) from sport-specific spatial-fidelity refinements, and characterizes a classification/spatial trade-off.
\end{enumerate}

\section{Related Work} \label{sec:related}

\paragraph{Stroke forecasting in racket sports}
Early tactical analysis used statistics and pattern mining over event sequences~\citep{patterndiscovery2010}. In badminton, CoachAI~\citep{Hsu2019CoachAI}, ShuttleSet~\citep{Wang2023ShuttleSet}, and a stroke-forecasting benchmark~\citep{Wang2024Benchmark} provide annotated data and evaluation settings. ShuttleNet~\citep{Wang2022ShuttleNet} combines rally progress with player styles to predict shot type and location; related work studies shot influence~\citep{Wang2021ShotInfluence}, movement~\citep{Chang2023DyMF}, and multimodal next-move prediction~\citep{Ibh2024}. Table tennis studies include video-based hit anticipation~\citep{Etaat2025LATTEMV} and racket-kinematics analysis~\citep{chou2025ttswing,cheng2025swing}. Seq2Event~\citep{Simpson2022Seq2Event} extends event-sequence prediction to soccer. Our focus is developing a multi-attribute prediction model across different racket sports by an explicit match hierarchy.
 
\paragraph{Transformers and feature-token modeling}
The Transformer~\citep{Vaswani2017} replaced recurrence with self-attention whose standard implementation has quadratic cost in sequence length. Long-sequence forecasters, e.g., Informer~\citep{Zhou2021Informer}, Autoformer~\citep{Wu2021Autoformer}, and TimesNet~\citep{Wu2023TimesNet}, improve efficiency or temporal structure. The iTransformer model~\citep{Liu2024iTransformer} attends across variables; we adopt this view, treating each semantic shot field as a token to capture interactions among shot attributes. In badminton, MuLMINet~\citep{Seong2023MuLMINet} couples a multi-input Transformer with a weighted loss. MT-HTA combines field-level interaction with explicit shot, rally, and higher-level context.
 
\paragraph{Hierarchical sequence models}
Hierarchical attention networks~\citep{Yang2016HAN} aggregate words and sentences, HLogformer~\citep{Hou2024HLogformer} represents nested log data, and hierarchical recurrent models analyze sentence purpose in academic abstracts~\citep{Hsu2021Academic}. MT-HTA applies this principle to competition units. Its shot--rally--set hierarchy, extended with games for tennis, differs from DyMF's hierarchical fusion of movement cues~\citep{Chang2023DyMF} and ShuttleNet's within-rally modeling~\citep{Wang2022ShuttleNet}.

\paragraph{Multi-task learning and loss balancing}
Multi-task learning shares representations to exploit inductive transfer across related tasks~\citep{Caruana1997}, which is well-suited to next-shot prediction, since several interrelated attributes jointly define a shot. Balancing competing tasks has been addressed by uncertainty weighting~\citep{Kendall2018} and gradient normalization~\citep{Chen2018GradNorm}, while architectural approaches route task-specific information through attention~\citep{Liu2019MTAN} or mixtures of experts~\citep{Ma2018MMoE}. MT-HTA uses a task-specific query over shared hierarchical context and evaluates random loss weighting~\citep{Lin2022RLW} as an alternative to fixed weights.

\section{Problem Formulation} \label{sec:formulation}
 
We represent a match as an ordered event sequence nested over a hierarchy of depth $D$. For the three-level case, a match $M$ contains $N_S$ top-level scoring units, each containing rallies and shots. We use \emph{set} for this unit following the table tennis and badminton annotations:
\begin{align}
M &= \{S_1,\dots,S_{N_S}\}, \label{eq:match}\\
S_i &= \{R^{(i)}_1,\dots,R^{(i)}_{N^{(i)}_R}\}, \label{eq:set}\\
R^{(i)}_j &= \{x_{i,j,1},\dots,x_{i,j,N^{(i,j)}_T}\}. \label{eq:rally}
\end{align}
For table tennis and badminton, the hierarchy is three levels (shot--rally--set); for tennis, it is four (shot--rally--game--set). Tennis inserts a game index between the set and rally indices in Eqs.~\eqref{eq:set}--\eqref{eq:rally}. Using table tennis as an example, each shot $x_{i,j,t}$ is a vector of semantic fields,
\begin{equation}
x_{i,j,t} = \big(\text{type},\,\text{loc},\,\text{bh},\,\text{spin},\,
\text{str},\,\text{sc}^{A},\,\text{sc}^{B},\,\text{pid},\,\text{oid},\,
\text{set}\big)_{i,j,t},
\label{eq:shotvec}
\end{equation}
comprising the predictable shot attributes (type, location, forehand/backhand, spin, strength), the score state of the two players ($\text{sc}^{A},\,\text{sc}^{B}$), their identities ($\text{pid},\,\text{oid}$), and the set index ($\text{set}$). The input-field set is dataset-specific: badminton additionally records player positions, while tennis includes game scores and serve-state fields.

We now consider a single target rally; with that rally fixed, the $i,j$ subscripts are constant, so we suppress them. For a prediction target within this rally, the model receives two inputs. The first is the \emph{current-rally prefix}---the shots already played in the ongoing rally, up to but excluding the target,
\begin{equation}
X_{1:t-1} = [x_1, x_2, \dots, x_{t-1}].
\label{eq:prefix}
\end{equation}
The second is the \emph{completed hierarchical context} $H_t$. For table tennis and badminton, $L_2$ contains completed rallies in the current set and $L_3$ contains completed sets. For tennis, $L_2$ contains completed rallies in the current game, $L_3$ contains completed games in the current set, and $L_4$ contains completed sets. Thus $L_1$ always contains only the current-rally prefix, and higher levels contain only units completed before the target shot.
 
Next-shot prediction is then the joint mapping
\begin{equation}
\big\{\hat{y}^{(k)}_{t}\big\}_{k\in\mathcal{T}}
   = f_\theta\big(X_{1:t-1};\,H_{t}\big),
   \qquad \hat{y}^{(k)}_t \in \mathbb{R}^{C_k},
\label{eq:target}
\end{equation}
where $\hat{y}^{(k)}_t$ denotes task logits, $\mathcal{T}$ is the dataset-specific task set, and $C_k$ is the corresponding output dimension, including any reserved padding class. The three sports have five, three, and four prediction tasks, respectively. Hierarchy depth, input fields, task vocabularies, and sequence-length limits are configured per dataset. Source-specific preprocessing maps records into this representation, after which the same model implementation is used with independently learned parameters $\theta$.

\section{Method: MT-HTA} \label{sec:method}
 
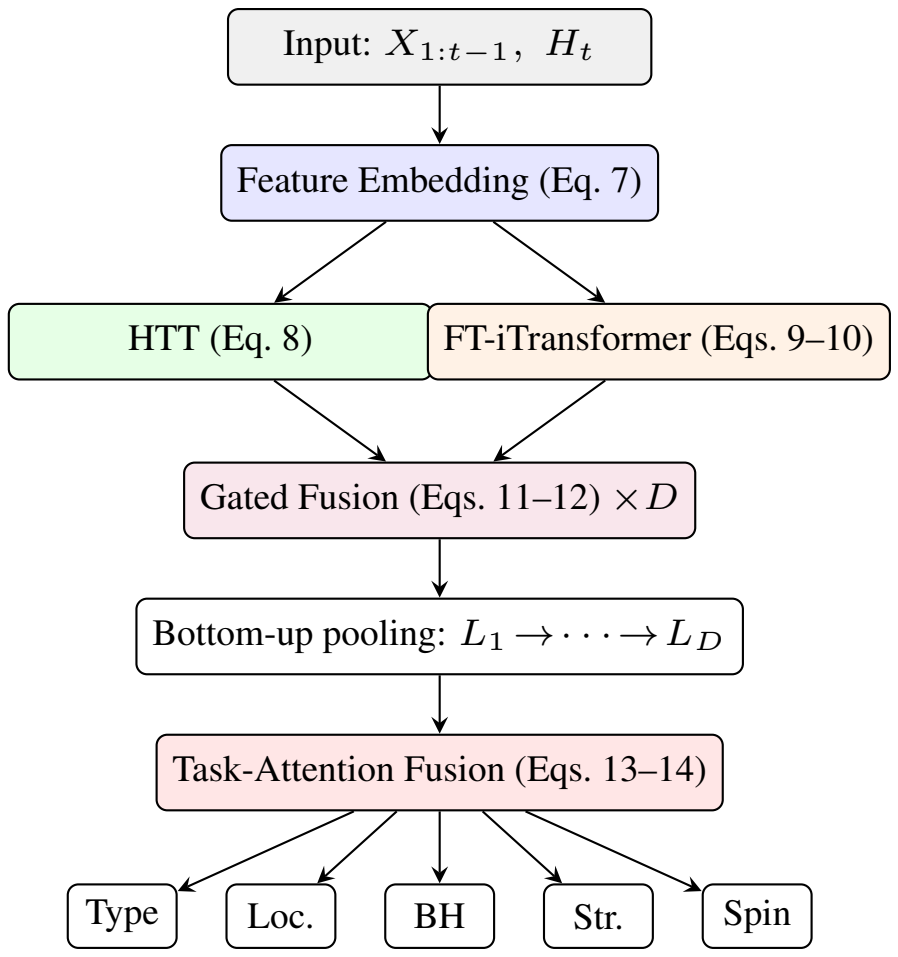
\begin{figure}[t]
\centering
\begin{tikzpicture}[
    box/.style={draw, rounded corners=2pt, align=center,
                minimum height=0.5cm, minimum width=2.8cm, font=\scriptsize,
                inner sep=3pt},
    sbox/.style={draw, rounded corners=2pt, align=center,
                 minimum height=0.42cm, minimum width=0.72cm, font=\scriptsize,
                 inner sep=2pt},
    arr/.style={->, >=stealth, thin},
]
\node[box, fill=gray!12] (inp) at (0,0) {Input: $X_{1:t-1},\;H_t$};
\node[box, fill=blue!10]  (emb) at (0,-0.9) {Feature Embedding (Eq.~\ref{eq:embed})};
\node[box, fill=green!10] (pact) at (-1.45,-1.95) {HTT (Eq.~\ref{eq:pact})};
\node[box, fill=orange!10](itf)  at ( 1.45,-1.95) {FT-iTransformer (Eqs.~\ref{eq:fttoken}--\ref{eq:itrans})};
\node[box, fill=purple!10](gate) at (0,-3.0) {Gated Fusion (Eqs.~\ref{eq:gate}--\ref{eq:fuse}) $\times D$};
\node[box]                (bu)   at (0,-3.9) {Bottom-up pooling: $L_1\!\to\!\cdots\!\to\!L_D$};
\node[box, fill=red!10]   (ta)   at (0,-4.8) {Task-Attention Fusion (Eqs.~\ref{eq:taskattn}--\ref{eq:taskffn})};
\node[sbox] (o1) at (-2.1,-5.75) {Type};
\node[sbox] (o2) at (-1.05,-5.75) {Loc.};
\node[sbox] (o3) at (0,-5.75)    {BH};
\node[sbox] (o4) at (1.05,-5.75) {Str.};
\node[sbox] (o5) at (2.1,-5.75)  {Spin};
\draw[arr] (inp)--(emb);
\draw[arr] (emb)--(pact);
\draw[arr] (emb)--(itf);
\draw[arr] (pact)--(gate);
\draw[arr] (itf)--(gate);
\draw[arr] (gate)--(bu);
\draw[arr] (bu)--(ta);
\foreach \o in {o1,o2,o3,o4,o5} { \draw[arr] (ta)--(\o); }
\end{tikzpicture}
\caption{Final MT-HTA overview. The hierarchical temporal Transformer (HTT) and FT-iTransformer encode each hierarchy level, with gated path fusion in the Final configuration. Task-specific queries attend to the resulting summaries and player context. The additional refinements are described in Section~\ref{sec:refine}.}
\label{fig:arch}
\end{figure}

Figure~\ref{fig:arch} summarizes the architecture. Each configured hierarchy level is encoded by a temporal path (hierarchical temporal Transformer, HTT) and a feature-token path (FT-iTransformer); the two paths are combined, levels are pooled bottom up, and per-task queries read the resulting summaries together with player context. The remainder of this section defines these components in order.
 
\subsection{Constructing the Prediction Context} \label{sec:construction}

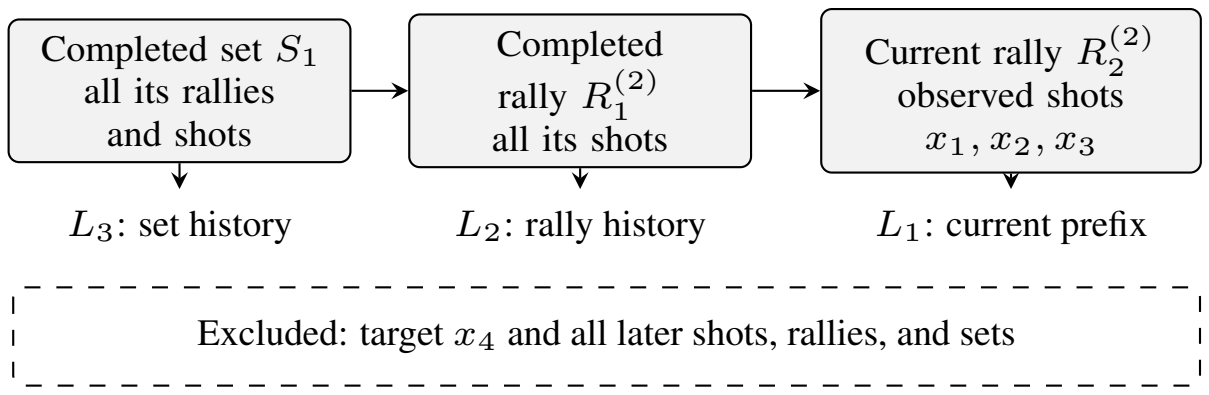
\begin{figure}[t]
\centering
\begin{tikzpicture}[
    x=1cm,y=1cm,
    every node/.style={font=\scriptsize,text=black,align=center},
    observed/.style={draw=black,rounded corners=2pt,
        fill=black!5,minimum height=0.85cm,inner sep=3pt},
    arr/.style={->,>=stealth,draw=black,thin}
]
\node[observed,text width=2.05cm] (set) at (1.1,0) {Completed set $S_1$\\all its rallies and shots};
\node[observed,text width=2.05cm] (rally) at (3.75,0) {Completed rally $R_1^{(2)}$\\all its shots};
\node[observed,text width=2.3cm] (prefix) at (6.6,0) {Current rally $R_2^{(2)}$\\observed shots $x_1,x_2,x_3$};
\draw[arr] (set) -- (rally);
\draw[arr] (rally) -- (prefix);
\node at (1.1,-0.9) {$L_3$: set history};
\node at (3.75,-0.9) {$L_2$: rally history};
\node at (6.6,-0.9) {$L_1$: current prefix};
\draw[arr] (set.south) -- (1.1,-0.65);
\draw[arr] (rally.south) -- (3.75,-0.65);
\draw[arr] (prefix.south) -- (6.6,-0.65);
\draw[draw=black,dashed,rounded corners=2pt]
    (0,-1.3) rectangle (7.85,-1.95);
\node at (3.925,-1.625) {Excluded: target $x_4$ and all later shots, rallies, and sets};
\end{tikzpicture}
\caption{Example of prefix-only input construction for a three-level match. Solid boxes contain observed events; the dashed target and future region is excluded. $R_j^{(i)}$ is rally $j$ in set $i$. The current rally contributes only its observed prefix.}
\label{fig:input-example}
\end{figure}

Figure~\ref{fig:input-example} makes the prediction boundary explicit. Consider the fourth shot of the second rally in the second set of a table tennis match. The current input $L_1$ contains exactly the first three shots of that rally. The completed first rally of the current set supplies $L_2$, while the completed first set supplies $L_3$. The target shot and the remaining events are not included in any level. Histories are grouped by their competition boundaries before they are encoded; a completed set is therefore not appended to the current-rally sequence.

At the start of the first rally, both higher-level histories are empty. After that rally ends, its observed events become eligible for the next rally's $L_2$ input. When a new set starts, $L_2$ is empty again, whereas the completed set becomes available to $L_3$. Missing history is represented by a learned empty-summary token, not by invented events. This absence is distinct from batch padding, whose positions are excluded by masks. Thus, the same prediction interface covers the opening, middle, set-boundary, and late-context groups evaluated in Section~\ref{sec:context}.

\subsection{Adapting the Event Schema}
The shared interface specifies field roles and nesting, while label dictionaries remain source-specific. Table tennis supplies Type, Backhand, Location, Strength, and Spin targets; badminton supplies Type, Backhand, and Landing Area; tennis supplies Type, Backhand, Location, and Spin. Badminton player-position fields are inputs rather than additional targets, and tennis adds game-score and serve-state inputs. Each source retains its own categorical vocabularies and output dimensions, e.g., a location category in one sport is not assumed to denote the same region in another. Only the configured tasks receive prediction heads and losses.

Tennis also changes the nesting: completed rallies belong to the current game ($L_2$), completed games to the current set ($L_3$), and completed sets to $L_4$. The temporal and feature-token operations below are reused at each configured level. This is architectural reuse with dataset-specific embeddings, task queries, and learned parameters, rather than joint training or parameter transfer between sports.

\paragraph{Adding a source}
Once records are converted into the shot--rally--higher-unit sample format, supporting a new dataset is a matter of declaration rather than of model code. A source specifies its hierarchy depth and level names, its categorical and numerical input fields with per-field embedding widths, the subset of fields that receive prediction heads, the per-task loss weights, and a sequence-length cap for each level. The encoder, path fusion, and task attention are untouched. The four corpora used here differ in these declarations and in the preprocessing that produces their samples, not in the model implementation.

\subsection{Hierarchy-Aligned Dual-Path Encoder} \label{sec:encoder}

\paragraph{Feature embedding}
Each recorded categorical field is mapped through its own embedding, and player and opponent identities pass through a shared player-style encoder to give style embeddings $e_p,e_o$. Numerical fields are kept as raw values. For the table tennis field set, the concatenation is projected to the model width $d_\text{model}$ as follows:
\begin{equation}
\begin{split}
c_t &= [\,e_p;\, e_o;\, e_\text{type};\, e_\text{loc};\, e_\text{bh};\,
         e_\text{spin};\, e_\text{str};\, \text{sc}^A;\, \text{sc}^B;\, \text{set}\,],\\
h^{(0)}_t &= \mathrm{LayerNorm}(W_\text{proj}\,c_t + b_\text{proj}).
\end{split}
\label{eq:embed}
\end{equation}
 
\paragraph{Temporal path}
The hierarchical temporal Transformer (HTT) adds sinusoidal positional encodings and applies $N_\text{layers}$ Transformer encoder layers along the time axis:
\begin{equation}
H^{(n)} = \mathrm{TransformerEncoderLayer}\big(H^{(n-1)}, \text{mask}\big),
\label{eq:pact}
\end{equation}
where $n=1,\dots,N_\text{layers}$, $H^{(0)}$ stacks the embedded vectors $h^{(0)}_t$ of one sequence, and the mask excludes padding positions. Causality is enforced by constructing inputs exclusively from the observed prefix and completed history; the encoder uses full self-attention within each observed sequence. Lower-level units are encoded separately before their summaries enter the next level.
 
\paragraph{Feature-interaction path (FT-iTransformer)}
In parallel, each categorical or numerical field forms a \emph{feature token}; player identities are handled separately by the style encoder. With $\mathcal{C}$ and $\mathcal{N}$ denoting the categorical and numerical input fields, $\mathcal{V}=\mathcal{C}\cup\mathcal{N}$ is indexed by $m$. For each $m\in\mathcal{V}$, the field's projected representations $\widetilde{U}^{(m)}_{\tau}\in\mathbb{R}^{d_\text{model}}$ at positions $\tau$ are right-aligned to the level-specific maximum length and padding is masked. They are projected along time with learnable, feature-specific coefficients $\alpha^{(m)}$,
\begin{equation}
z^{(m)} = \sum_{\tau=1}^{T^{(\ell)}_\text{max}} \alpha^{(m)}_\tau\,
\widetilde{U}^{(m)}_{\tau},
\label{eq:fttoken}
\end{equation}
where $T^{(\ell)}_\text{max}$ is the maximum sequence length at hierarchy level $\ell$, yielding one token per field. Stacking the $F=|\mathcal{V}|$ tokens into $Z_\text{feat}=[\,z^{(1)};\dots;z^{(F)}\,]\in\mathbb{R}^{F\times d_\text{model}}$, self-attention runs \emph{across fields}, and a mean over feature tokens gives the path summary (computed at each hierarchy level):
\begin{equation}
\begin{split}
\widetilde{Z} &= \mathrm{TransformerEncoder}\!\bigl(\mathrm{LayerNorm}(Z_\text{feat})\bigr),\\
h_\text{iTrans} &= \mathrm{LayerNorm}\!\bigl(\tfrac{1}{F}\textstyle\sum_{m=1}^{F}\widetilde{Z}_{m,:}\bigr).
\end{split}
\label{eq:itrans}
\end{equation}
At higher levels, field summaries retain their semantics: categorical embeddings are averaged over valid lower-level units, while numerical state fields retain the last valid value. HTT models temporal progression, and the feature-token path models interactions among these fields.
 
\paragraph{Path fusion}
MT-HTA Core concatenates the two path summaries at each level, projects them to $d_\text{model}$, and applies layer normalization. Final MT-HTA replaces this projection with sigmoid-gated fusion. With $h^{(\ell)}_\text{HTT}$ pooled from the temporal path and $h^{(\ell)}_\text{iTrans}$ from the feature-token path, and $\odot$ denoting the elementwise product, Final uses
\begin{align}
g^{(\ell)} &= \sigma\!\big(W_g^\text{fuse}\,[\,h^{(\ell)}_\text{HTT}; h^{(\ell)}_\text{iTrans}\,] + b_g^\text{fuse}\big),
\label{eq:gate}\\
h^{(\ell)}_\text{fused} &= \mathrm{LayerNorm}\!\big(g^{(\ell)}\odot h^{(\ell)}_\text{HTT}
   + (1-g^{(\ell)})\odot h^{(\ell)}_\text{iTrans}\big).
\label{eq:fuse}
\end{align}
 
\paragraph{Bottom-up flow}
Temporal encoding proceeds shot $\rightarrow$ rally $\rightarrow$ set, or shot $\rightarrow$ rally $\rightarrow$ game $\rightarrow$ set for tennis. Last-valid-token pooling produces lower-level summaries for the next level's sequence. The feature-token path aggregates fields separately as described above. A level with no completed history uses a learned empty-summary token.
 
\subsection{Task-Attention Fusion} \label{sec:taskattn}

For the three-level case, hierarchical summaries and player embeddings form $S_\text{ctx} = [\,h_\text{shot}; h_\text{rally}; h_\text{set}; e'_p; e'_o\,]$; tennis inserts $h_\text{game}$ before $h_\text{set}$. Here $e'_p,e'_o$ are style embeddings projected to the model width. Each task $k$ has a learnable query $q_k$ that attends over this context and is refined by a task-specific feed-forward block:
\begin{align}
u^{(k)}_t &= \mathrm{MultiHeadAttn}\big(Q=q_k,\,K=S_\text{ctx},\,V=S_\text{ctx}\big),
\label{eq:taskattn}\\
z^{(k)}_t &= \mathrm{LayerNorm}\big(u^{(k)}_t + \mathrm{FFN}_k(u^{(k)}_t)\big).
\label{eq:taskffn}
\end{align}
The per-task vector is decoded by a linear head, $\hat{y}^{(k)}_{t} = W^{(k)}_\text{out}\,z^{(k)}_t + b^{(k)}_\text{out}$. Independent queries allow tasks to weight the shared context differently. The encoder is optimized jointly.
 
\subsection{Domain-Informed Refinements} \label{sec:refine}

In addition to gated path fusion, Final MT-HTA enables four refinements. \emph{Shot-aware positional encoding} adds a serve flag and a self/opponent flag to the shot input to represent rally starts and player identity within the exchange:
\begin{equation}
h^{(\text{SA})}_t = h^{(0)}_t + \mathrm{Emb}_\text{serve}(\text{is\_serve}_t)
   + \mathrm{Emb}_\text{self}(\text{is\_self}_t).
\label{eq:shotaware}
\end{equation}
\emph{Top-down refinement} lets the highest available summary $h_\text{high}$ query the full shot sequence $H_\text{shot}$, the $L_1$ output of Eq.~\ref{eq:pact} before pooling, and feeds the result back with a learnable residual scale:
\begin{align}
h_\text{TD} &= \mathrm{MultiHeadAttn}\big(Q=h_\text{high},\,K=H_\text{shot},\,V=H_\text{shot}\big),
\label{eq:topdown}\\
\tilde{h}_\text{shot} &= h_\text{shot} + \gamma_\text{TD}\,h_\text{TD}.
\label{eq:topdownres}
\end{align}
\emph{Gated skip window} re-injects the most recent shot ($w=1$), projected to $s_\text{proj}$, into each task representation through a sigmoid gate, preserving the local cue that deep aggregation can wash out:
\begin{align}
g^{(k)} &= \sigma\!\big(W_g^\text{skip}[\,z^{(k)}_t;\, s_\text{proj}\,] + b_g^\text{skip}\big), \label{eq:skipg}\\
\tilde{z}^{(k)}_t &= g^{(k)}\odot z^{(k)}_t + (1-g^{(k)})\odot s_\text{proj}. \label{eq:skip}
\end{align}
Finally, \emph{turn-based gating} hard-swaps the player and opponent style embeddings at each step according to who strikes, keeping the context representation hitter-centric across alternating turns. All five optional components are enabled during training of Final MT-HTA. Their effects are evaluated by retraining ablated configurations.
 
\subsection{Training Objective} \label{sec:loss}
All tasks use label-smoothed cross-entropy ($\epsilon=0.1$). For table tennis location, each half of the table is divided into a $3\times3$ grid of nine cells, and a tenth, non-grid category covers all other outcomes (e.g., out of bounds); we additionally penalize spatial error on the nine grid cells. Let $\mathcal{B}_G$ denote batch examples whose target lies in the grid, and let $\hat{p}^{(n,G)}$ be the softmax restricted to the nine grid logits. Then
\begin{align}
\mathcal{L}_\text{dist} &= \frac{1}{|\mathcal{B}_G|}\sum_{n\in\mathcal{B}_G}\sum_{i=1}^{9}
   \hat{p}^{(n,G)}_i\, \mathrm{Dist}_{i,\,y^{(n)}}, \label{eq:ldist}\\
\mathcal{L}_\text{location} &= \mathcal{L}^{(\text{loc})}_\text{CE}
   + \lambda_\text{dist}\,\mathcal{L}_\text{dist}. \label{eq:dist}
\end{align}
Here $\mathrm{Dist}$ is Euclidean distance in grid-cell units, $\lambda_\text{dist}=1.0$, and the distance term is zero when $\mathcal{B}_G$ is empty. Cross-entropy retains the non-grid category and ignores padding. The badminton and tennis configurations use cross-entropy without this grid-distance term. The total loss is $\mathcal{L}_\text{total}=\sum_{k\in\mathcal{T}}w_k\mathcal{L}_k$, with fixed weights: $0.2$ per task for table tennis, $0.25$ for tennis, and $0.34/0.33/0.33$ for badminton Type/Backhand/Location. Random loss weighting was also evaluated, but did not improve Final MT-HTA's mean Macro F1 on original table tennis and is excluded from the reported Final configuration.

\section{Experiments} \label{sec:exp}

\begin{table*}[t]
\centering
\caption{Dataset overview. $D$ is hierarchy depth. Class counts are in task order: Type / Backhand / Location for every dataset, plus Strength / Spin for table tennis and Spin for tennis. These are semantic annotation classes; counts exclude the model's index-$0$ padding class, so each output dimension $C_k$ is one larger. Table tennis location divides each half of the table into a $3\times3$ grid plus a non-grid category (e.g., out of bounds); location definitions differ across sports.}
\label{tab:datasets}
\resizebox{\textwidth}{!}{
\begin{tabular}{lcccccccl}
\toprule
Dataset & $D$ & Matches & Rallies & Shots & Players & Tasks & Classes per task & Annotation source \\
\midrule
Table Tennis (orig.) & 3 & 116 & 9{,}303 & 45{,}714 & 82 & 5 & 19 / 3 / 10 / 4 / 6 & technical--tactical system \\
Table Tennis (exp.)  & 3 & 271 & 20{,}273 & 108{,}289 & 189 & 5 & 19 / 3 / 10 / 4 / 6 & technical--tactical system \\
Badminton (ShuttleSet) & 3 & 94 & 5{,}953 & 64{,}378 & 43 & 3 & 10 / 2 / 10 & expert-annotated, public \\
Tennis (MCP) & 4 & 100 & 15{,}778 & 75{,}937 & 78 & 4 & 11 / 2 / 7 / 3 & crowdsourced, public \\
\bottomrule
\end{tabular}
}
\end{table*}

\paragraph{Datasets and setup}
We evaluate on four datasets spanning three racket sports (Table~\ref{tab:datasets}). Table tennis records come from a technical--tactical annotation system previously used for stroke-type prediction~\citep{cai2025stroke}; the expanded version adds matches and a broader player population. The badminton dataset combines ShuttleSet and ShuttleSet22~\citep{Wang2023ShuttleSet,Wang2024Benchmark}, and tennis uses converted Match Charting Project records~\citep{Sackmann_MCP}. Dataset-specific preprocessing preserves competition boundaries and maps categorical fields to the model's vocabularies.

We use a fixed $70/10/20$ train/validation/test split by match, preventing a match from appearing in multiple partitions but does not require disjoint players or pairings. Learned-model results are means and sample standard deviations over three training trials. For each task, F1 is weighted by class support; ``Macro F1'' denotes the arithmetic mean of these task-level weighted F1 scores. For table tennis, Location Average Distance is the mean Euclidean distance in grid-cell units over examples where both the target and the argmax prediction lie in the valid $3\times3$ grid, so it must be read alongside Location F1. Throughout, bold marks the best mean in a comparison but may not indicate a statistically significant difference.

\paragraph{Implementation}
The compared models use $d_\text{model}=256$, which standardizes hidden width but does not equalize parameter counts. Training uses AdamW (learning rate $10^{-4}$, weight decay $0.01$) for 40 epochs, a cosine schedule with $10\%$ warm-up, and gradient clipping at $1.0$. We evaluate the checkpoint with the lowest validation loss. The accompanying configurations specify batch size 128 for table tennis and badminton and 64 for tennis; the task weights and distance-loss settings are given in Section~\ref{sec:loss}. Training and the reported evaluations use a single NVIDIA V100 GPU ($32$\,GB), 4 vCPUs, and $60$\,GB RAM, with Python 3.9 and PyTorch 2.3.0 on Linux. Inference is a single forward pass and does not use this GPU; the millisecond-scale latency noted in Section~\ref{sec:intro} was measured separately on a laptop CPU (Apple~M2, four threads).

\paragraph{Comparison models and input scope}
Table~\ref{tab:baseline-designs} distinguishes access to history from the way that history is encoded. Current BiLSTM and adapted ShuttleNet use only the current-rally prefix. Rally-context LSTM adds a mean-pooled representation of completed rallies, while Hierarchical LSTM explicitly models shot, rally, and set sequences. Flat Transformer receives completed history as well as the current prefix, but concatenates them into one shot sequence. Thus, comparison with the flat model is not a comparison against a current-rally-only input.

\begin{table}[t]
\centering
\caption{Compared models on original table tennis: the history each model receives and how it encodes that history. $L_1$ is the current-rally prefix, $L_2$ the completed rallies in the current set, and $L_3$ the completed sets; ``Full'' includes all three when available, within the configured input limits. All inputs precede the target.}
\label{tab:baseline-designs}
\setlength{\tabcolsep}{3pt}
\begin{tabular}{@{}lll@{}}
\toprule
Model & History & Representation \\
\midrule
Current BiLSTM & $L_1$ & Shot recurrence \\
ShuttleNet (adapted) & $L_1$ & Encoder--decoder \\
Rally-context LSTM & $L_1+L_2$ & LSTM + history mean \\
Hierarchical LSTM & Full & Hierarchical recurrence \\
Flat Transformer & Full & Flat shot tokens \\
PatchTST (adapted) & Full & Per-field temporal patches \\
FT-iTransformer & Full & Flat feature tokens \\
MT-HTA Core / Final & Full & Hierarchical dual paths \\
\bottomrule
\end{tabular}
\end{table}

We also report an adapted PatchTST~\citep{Nie2023PatchTST} baseline in the same original table tennis comparison (Table~\ref{tab:baseline}). Each categorical field is embedded and each numerical field is projected before overlapping temporal patches are formed within each semantic channel. A shared Transformer encodes channels independently; pooled channel summaries are subsequently fused for multi-task classification. This differs from FT-iTransformer, which attends across field tokens, and from MT-HTA, which additionally retains the competition hierarchy.
 
\subsection{Cross-Sport Main Results}
\label{sec:main}

\begin{table*}[t]
\centering
\caption{Cross-sport main results (Macro F1); learned models report mean$\pm$std over three trials.}
\label{tab:main}
\begin{tabular}{lccccc}
\toprule
Dataset & Majority & Rally-ctx LSTM & Flat Transformer & MT-HTA Core & Final MT-HTA \\
\midrule
Table Tennis (orig.)  & 0.2532 & $0.5177_{\pm.0030}$ & $0.5191_{\pm.0059}$ & $0.5302_{\pm.0065}$ & $\mathbf{0.5315_{\pm.0047}}$ \\
Table Tennis (exp.)   & 0.2379 & $0.5405_{\pm.0057}$ & $0.5500_{\pm.0011}$ & $0.5474_{\pm.0013}$ & $\mathbf{0.5525_{\pm.0007}}$ \\
Badminton (ShuttleSet)& 0.2175 & $0.5308_{\pm.0028}$ & $0.5238_{\pm.0041}$ & $0.5289_{\pm.0070}$ & $\mathbf{0.5309_{\pm.0028}}$ \\
Tennis (MCP)          & 0.5349 & $0.7038_{\pm.0049}$ & $0.6969_{\pm.0012}$ & $0.7074_{\pm.0014}$ & $\mathbf{0.7095_{\pm.0064}}$ \\
\bottomrule
\end{tabular}
\end{table*}

Final MT-HTA attains the highest mean Macro F1 on every dataset in Table~\ref{tab:main}. Which generic baseline comes closest, however, changes from sport to sport: the hierarchical LSTM is the strongest on original table tennis (Table~\ref{tab:baseline}), whereas the rally-context LSTM is the strongest on badminton and tennis. Ranking first on all four therefore does not reduce to outperforming a single fixed competitor.

MT-HTA Core is close to Final. However, Core trails the flat Transformer on expanded table tennis and the rally-context LSTM on badminton. These results support reuse of the architecture across the evaluated datasets, without establishing uniform superiority or cross-sport transfer.

The Majority column in Table~\ref{tab:main} calibrates these scores across sports: tennis starts from a far higher constant-prediction baseline, so its higher absolute Macro F1 partly reflects a more concentrated label distribution rather than stronger prediction. Macro F1 remains support-weighted throughout, so none of these comparisons speaks to performance on rare labels.

\subsection{Task-Level Differences across Sports}
Table~\ref{tab:crosssport-tasks} expands the aggregate comparison into the tasks available for each dataset. Scores are compared within each panel: Type and Location use different vocabularies across sports, and an absent task is not assigned a zero score. The table exposes differences that can be hidden by averaging over tasks.

\begin{table*}[t]
\centering
\caption{Task-level cross-sport results (mean$\pm$std). Each F1 is support-weighted. Bold marks the highest mean F1 or lowest mean distance among the four models within a dataset. A dash marks an attribute the dataset does not annotate or, in the Loc.\ Dist.\ column, a metric computed only for table tennis on its $3\times3$ grid. For badminton, Location F1 reports Landing Area.}
\label{tab:crosssport-tasks}
\resizebox{\textwidth}{!}{%
\begin{tabular}{lcccccc}
\toprule
Model & Type F1 & Backhand F1 & Location F1 & Strength F1 & Spin F1 & Loc.\ Dist.\ $\downarrow$ \\
\midrule
\multicolumn{7}{l}{\textit{(a) Expanded table tennis}} \\
Rally-context LSTM & $0.4773_{\pm.0076}$ & $0.7756_{\pm.0011}$ & $0.3096_{\pm.0079}$ & $0.5381_{\pm.0151}$ & $0.6018_{\pm.0023}$ & $0.8562_{\pm.0115}$ \\
Flat Transformer & $0.4914_{\pm.0054}$ & $0.7731_{\pm.0024}$ & $\mathbf{0.3172_{\pm.0047}}$ & $0.5577_{\pm.0037}$ & $0.6105_{\pm.0038}$ & $\mathbf{0.8315_{\pm.0040}}$ \\
MT-HTA Core & $0.4884_{\pm.0053}$ & $0.7778_{\pm.0009}$ & $0.3064_{\pm.0045}$ & $\mathbf{0.5593_{\pm.0118}}$ & $0.6051_{\pm.0007}$ & $0.8572_{\pm.0060}$ \\
Final MT-HTA & $\mathbf{0.5009_{\pm.0015}}$ & $\mathbf{0.7792_{\pm.0043}}$ & $0.3104_{\pm.0007}$ & $0.5565_{\pm.0039}$ & $\mathbf{0.6156_{\pm.0025}}$ & $0.8567_{\pm.0043}$ \\
\midrule
\multicolumn{7}{l}{\textit{(b) Badminton}} \\
Rally-context LSTM & $0.5002_{\pm.0088}$ & $\mathbf{0.8333_{\pm.0030}}$ & $0.2590_{\pm.0030}$ & -- & -- & -- \\
Flat Transformer & $0.4890_{\pm.0114}$ & $0.8245_{\pm.0021}$ & $0.2578_{\pm.0030}$ & -- & -- & -- \\
MT-HTA Core & $0.5020_{\pm.0093}$ & $0.8301_{\pm.0065}$ & $0.2545_{\pm.0067}$ & -- & -- & -- \\
Final MT-HTA & $\mathbf{0.5072_{\pm.0039}}$ & $0.8236_{\pm.0044}$ & $\mathbf{0.2620_{\pm.0037}}$ & -- & -- & -- \\
\midrule
\multicolumn{7}{l}{\textit{(c) Tennis}} \\
Rally-context LSTM & $0.7856_{\pm.0065}$ & $0.7484_{\pm.0063}$ & $0.4559_{\pm.0038}$ & -- & $0.8252_{\pm.0036}$ & -- \\
Flat Transformer & $0.7753_{\pm.0027}$ & $0.7469_{\pm.0041}$ & $0.4568_{\pm.0014}$ & -- & $0.8085_{\pm.0017}$ & -- \\
MT-HTA Core & $0.7889_{\pm.0014}$ & $0.7541_{\pm.0115}$ & $\mathbf{0.4607_{\pm.0039}}$ & -- & $0.8259_{\pm.0036}$ & -- \\
Final MT-HTA & $\mathbf{0.7897_{\pm.0010}}$ & $\mathbf{0.7640_{\pm.0188}}$ & $0.4584_{\pm.0036}$ & -- & $\mathbf{0.8260_{\pm.0024}}$ & -- \\
\bottomrule
\end{tabular}%
}
\end{table*}

On expanded table tennis, Final has the highest mean Type, Backhand, and Spin F1, while Core has the highest Strength F1. Flat Transformer remains best on Location F1 ($0.3172$ vs.\ Final's $0.3104$) and Location Distance ($0.8315$ vs.\ $0.8567$). Its spatial advantage is therefore not visible in Macro F1 alone. The distance comparison also retains the valid-prediction restriction described above; it is not an unconditional error over every test example.

On badminton, Final's Type and Landing Area means exceed those of Rally-context LSTM ($0.5072$ vs.\ $0.5002$, and $0.2620$ vs.\ $0.2590$), while its Backhand mean is lower ($0.8236$ vs.\ $0.8333$). These opposing differences explain why their Macro F1 scores are nearly equal.

On tennis, Final leads in mean Type, Backhand, and Spin F1, but Core has the highest Location F1 ($0.4607$ vs.\ Final's $0.4584$). The largest Final--Core task-level gap is in Backhand, whose Final score also has substantial seed variation ($0.7640\pm0.0188$). Across the three panels, an overall improvement need not benefit every output, and the optional refinements do not have a uniform effect across sports.

\subsection{Component Analysis} \label{sec:component}
 
\begin{table*}[t]
\centering
\caption{Baseline comparison on original table tennis (three-seed mean$\pm$std, $d_\text{model}=256$). Bold = column-wise best. ShuttleNet and PatchTST are adapted for single-step multi-task classification; Table~\ref{tab:baseline-designs} specifies their input scope.}
\label{tab:baseline}
\small
\setlength{\tabcolsep}{3pt}
\resizebox{\textwidth}{!}{%
\begin{tabular}{lccccccc}
\toprule
Setting & Macro F1 & Type F1 & Backhand F1 & Location F1 & Strength F1 & Spin F1 & Loc.\ Dist.\ $\downarrow$ \\
\midrule
ShuttleNet (Seq2Seq)   & $0.3958_{\pm.0260}$ & $0.2672_{\pm.0439}$ & $0.5690_{\pm.0411}$ & $0.2046_{\pm.0062}$ & $0.4376_{\pm.0015}$ & $0.5009_{\pm.0423}$ & $0.9607_{\pm.0061}$ \\
Current BiLSTM         & $0.5154_{\pm.0034}$ & $0.4578_{\pm.0054}$ & $0.7417_{\pm.0028}$ & $0.2550_{\pm.0123}$ & $0.5101_{\pm.0018}$ & $0.6124_{\pm.0021}$ & $0.8919_{\pm.0080}$ \\
Rally-context LSTM     & $0.5177_{\pm.0030}$ & $0.4644_{\pm.0140}$ & $0.7439_{\pm.0012}$ & $0.2666_{\pm.0057}$ & $0.5036_{\pm.0040}$ & $0.6101_{\pm.0109}$ & $0.9041_{\pm.0144}$ \\
Hierarchical LSTM      & $0.5206_{\pm.0009}$ & $0.4715_{\pm.0045}$ & $0.7443_{\pm.0023}$ & $0.2627_{\pm.0098}$ & $0.5063_{\pm.0045}$ & $0.6180_{\pm.0081}$ & $0.9001_{\pm.0145}$ \\
Flat Transformer       & $0.5191_{\pm.0059}$ & $0.4598_{\pm.0090}$ & $0.7355_{\pm.0024}$ & $0.2667_{\pm.0122}$ & $0.5148_{\pm.0041}$ & $0.6185_{\pm.0060}$ & $\mathbf{0.8791_{\pm.0012}}$ \\
PatchTST (adapted) & $0.4635_{\pm.0091}$ & $0.3648_{\pm.0121}$ & $0.6598_{\pm.0265}$ & $0.2373_{\pm.0148}$ & $0.4855_{\pm.0045}$ & $0.5700_{\pm.0056}$ & $0.9157_{\pm.0174}$ \\
\midrule
MT-HTA Core            & $0.5302_{\pm.0065}$ & $0.4848_{\pm.0036}$ & $\mathbf{0.7457_{\pm.0004}}$ & $\mathbf{0.2693_{\pm.0043}}$ & $0.5273_{\pm.0094}$ & $0.6238_{\pm.0176}$ & $0.9046_{\pm.0055}$ \\
Full Final MT-HTA      & $\mathbf{0.5315_{\pm.0047}}$ & $\mathbf{0.4891_{\pm.0069}}$ & $0.7429_{\pm.0061}$ & $0.2686_{\pm.0060}$ & $\mathbf{0.5310_{\pm.0049}}$ & $\mathbf{0.6259_{\pm.0041}}$ & $0.8874_{\pm.0101}$ \\
\bottomrule
\end{tabular}
}
\end{table*}

\begin{table}[t]
\centering
\caption{Hierarchy depth and task-fusion variants relative to MT-HTA Core on original table tennis (three-seed mean$\pm$std, $d_\text{model}=256$). All rows disable the optional components. L1, L1+L2, and Core use Task Attention (TA); the other variants use all three levels with different task-fusion operators. Bold = column-wise best within this table.}
\label{tab:fusion}
\begin{tabular}{lccc}
\toprule
Setting & Macro F1 & Location F1 & Loc. Dist. $\downarrow$ \\
\midrule
L1 only (TA) & $0.5204_{\pm.0054}$ & $0.2677_{\pm.0078}$ & $\mathbf{0.8902_{\pm.0059}}$ \\
L1+L2 (TA) & $0.5250_{\pm.0042}$ & $0.2677_{\pm.0009}$ & $0.8944_{\pm.0187}$ \\
Task Projection & $0.5271_{\pm.0048}$ & $0.2643_{\pm.0019}$ & $0.9086_{\pm.0190}$ \\
CLS-token Fusion & $0.5278_{\pm.0025}$ & $0.2634_{\pm.0078}$ & $0.9104_{\pm.0142}$ \\
MT-HTA Core & $\mathbf{0.5302_{\pm.0065}}$ & $\mathbf{0.2693_{\pm.0043}}$ & $0.9046_{\pm.0055}$ \\
\bottomrule
\end{tabular}
\end{table}
 
On original table tennis, both Core and Final have higher mean Macro F1 than the recurrent and flat Transformer baselines (Table~\ref{tab:baseline}). The flat Transformer has the lowest Location Distance, indicating that aggregate classification and spatial error can favor different models.

Adapted PatchTST reaches $0.4635\pm0.0091$ Macro F1, below Flat Transformer's $0.5191\pm0.0059$. Both use flattened available history, but organize it into different tokens and encoders. This comparison shows that access to longer history through temporal patches does not by itself match the other encoders in this setup. It does not isolate patching, channel independence, or the categorical-task adaptation as the cause of the gap.
 
Table~\ref{tab:fusion} examines hierarchy depth and the fusion of hierarchical summaries into task representations.

\begin{table*}[t]
\centering
\caption{Module sensitivity on original table tennis (three-seed mean$\pm$std). Each optional component is disabled in turn relative to Full Final MT-HTA, and each ablated model is retrained; ``w/o Gated Path Fusion'' restores concatenation and projection. MT-HTA Core disables all five components and is repeated from Table~\ref{tab:fusion}. Bold = column-wise best within this table.}
\label{tab:ablation}
\small
\setlength{\tabcolsep}{3pt}
\resizebox{\textwidth}{!}{%
\begin{tabular}{lccccccc}
\toprule
Setting & Macro F1 & Type F1 & Backhand F1 & Location F1 & Strength F1 & Spin F1 & Loc.\ Dist.\ $\downarrow$ \\
\midrule
Full Final MT-HTA      & $0.5315_{\pm.0047}$ & $0.4891_{\pm.0069}$ & $0.7429_{\pm.0061}$ & $0.2686_{\pm.0060}$ & $0.5310_{\pm.0049}$ & $0.6259_{\pm.0041}$ & $\mathbf{0.8874_{\pm.0101}}$ \\
MT-HTA Core            & $0.5302_{\pm.0065}$ & $0.4848_{\pm.0036}$ & $0.7457_{\pm.0004}$ & $0.2693_{\pm.0043}$ & $0.5273_{\pm.0094}$ & $0.6238_{\pm.0176}$ & $0.9046_{\pm.0055}$ \\
\midrule
\;\;w/o Shot-Aware PE       & $0.5296_{\pm.0017}$ & $0.4800_{\pm.0122}$ & $0.7432_{\pm.0032}$ & $0.2719_{\pm.0049}$ & $0.5245_{\pm.0048}$ & $0.6282_{\pm.0037}$ & $0.9075_{\pm.0067}$ \\
\;\;w/o Gated Path Fusion  & $0.5301_{\pm.0041}$ & $0.4850_{\pm.0062}$ & $0.7461_{\pm.0060}$ & $0.2700_{\pm.0006}$ & $0.5273_{\pm.0064}$ & $0.6221_{\pm.0048}$ & $0.8891_{\pm.0090}$ \\
\;\;w/o Top-Down Attention  & $0.5272_{\pm.0039}$ & $0.4804_{\pm.0029}$ & $0.7431_{\pm.0038}$ & $0.2622_{\pm.0035}$ & $0.5274_{\pm.0049}$ & $0.6231_{\pm.0085}$ & $0.9091_{\pm.0216}$ \\
\;\;w/o Turn-Based Gating   & $0.5330_{\pm.0018}$ & $0.4846_{\pm.0051}$ & $0.7455_{\pm.0035}$ & $0.2722_{\pm.0030}$ & $\mathbf{0.5342_{\pm.0033}}$ & $0.6286_{\pm.0034}$ & $0.8940_{\pm.0178}$ \\
\;\;w/o Skip Window         & $\mathbf{0.5345_{\pm.0016}}$ & $\mathbf{0.4897_{\pm.0025}}$ & $\mathbf{0.7482_{\pm.0024}}$ & $\mathbf{0.2744_{\pm.0132}}$ & $0.5307_{\pm.0044}$ & $\mathbf{0.6298_{\pm.0057}}$ & $0.9142_{\pm.0115}$ \\
\bottomrule
\end{tabular}
}
\end{table*}

With Task Attention, mean Macro F1 increases from $0.5204$ for L1 to $0.5250$ for L1+L2 and $0.5302$ for all three levels (Table~\ref{tab:fusion}). At that depth, Task Attention also has a higher mean than CLS-token fusion ($0.5278$) and Task Projection ($0.5271$). These differences are modest relative to seed variability. Location Distance follows the opposite mean trend across the depth variants. Thus, the observed classification gains are accompanied by higher spatial error in this comparison.

Final reduces Location Distance from Core's $0.9046$ to $0.8874$ (Table~\ref{tab:ablation}), while the flat Transformer remains lower at $0.8791$ (Table~\ref{tab:baseline}). Removing the skip window, top-down attention, or shot-aware positional encoding from Final increases mean distance by $0.0268$, $0.0217$, and $0.0201$, respectively. Removing turn-based gating or gated path fusion increases it by smaller amounts, $0.0066$ and $0.0017$. Conversely, the configurations without the skip window or turn-based gating have higher mean Macro F1 than Final ($0.5345$ and $0.5330$ vs.\ $0.5315$).
 
The ablations therefore describe a classification--spatial trade-off on original table tennis. Each ablation reflects a component's effect within the full configuration.

\begin{table}[t]
\centering
\caption{Dual-path analysis on original table tennis (three-seed mean$\pm$std). The standalone FT-iTransformer baseline and Final without the feature-token path are compared with Final.}
\label{tab:dualpath}
\begin{tabular}{lcc}
\toprule
Setting & Macro F1 & Loc.\ Dist.\ $\downarrow$ \\
\midrule
FT-iTransformer only        & $0.5098_{\pm.0037}$ & $0.8995_{\pm.0077}$ \\
Final w/o FT-iTransformer   & $0.5280_{\pm.0051}$ & $0.9153_{\pm.0124}$ \\
Final MT-HTA                & $\mathbf{0.5315_{\pm.0047}}$ & $\mathbf{0.8874_{\pm.0101}}$ \\
\bottomrule
\end{tabular}
\end{table}

The standalone FT-iTransformer reaches $0.5098$ Macro F1 (Table~\ref{tab:dualpath}). Removing the feature-token path from Final lowers mean Macro F1 by $0.0035$ and raises Location Distance from $0.8874$ to $0.9153$. Type and Strength F1 also decrease, while Backhand and Location F1 increase slightly. This ablation supports the path's complementary role within the full model, with task-dependent effects.
 
\subsection{Expanded Data and Multi-Step Prediction} \label{sec:robust}
 
The expanded table tennis dataset increases the number of matches from $116$ to $271$, players from $82$ to $189$, and shots from $45{,}714$ to $108{,}289$ (Table~\ref{tab:datasets}). Final achieves $0.5525$ Macro F1, compared with $0.5500$ for the flat Transformer and $0.5474$ for Core (Table~\ref{tab:main}). Final's margin over the flat Transformer is smaller than on original table tennis ($0.0025$ vs.\ $0.0124$), and Core no longer leads that baseline.

\paragraph{Autoregressive evaluation protocol}
To examine feedback from earlier predictions, we evaluate autoregressive Type prediction on original table tennis using the three trained Final checkpoints. The model observes $n=1,\ldots,5$ initial shots and predicts up to five further steps, provided the targets exist. Each appended event uses the argmax labels for all five tasks, although this diagnostic scores only Type. Completed-rally and completed-set histories remain observed context.

Player identities, score state, and other non-target fields retain their recorded values. Rollout stops at the recorded rally boundary, so this is conditional label prediction, not autonomous match simulation. Top-3 measures membership in a ranked candidate set; only argmax labels are fed back, without branching into alternative trajectories.

\begin{table}[t]
\centering
\caption{Autoregressive Type accuracy on original table tennis (three-seed mean$\pm$std). The model observes the first $n\in\{1,\ldots,5\}$ shots of a rally and predicts onward, feeding back its own argmax predictions; Step $+k$ scores the $k$-th predicted shot, pooling all prefix lengths $n$ by sample count. Each pair is one rally prefix evaluated at one step, and Pairs counts them; the count depends only on the fixed test set and is identical across seeds. Top-1 accuracy requires the argmax Type to be correct, and Top-3 accuracy requires the true Type to be among the three highest-scoring classes. Overall pools all steps.}
\label{tab:rollout}
\begin{tabular}{@{}lrcc@{}}
\toprule
Step & Pairs & Top-1 accuracy & Top-3 accuracy \\
\midrule
$+1$ & $6{,}602$ & $0.5146_{\pm.0061}$ & $0.8501_{\pm.0049}$ \\
$+2$ & $5{,}070$ & $0.3142_{\pm.0075}$ & $0.6151_{\pm.0024}$ \\
$+3$ & $3{,}638$ & $0.2623_{\pm.0126}$ & $0.5456_{\pm.0122}$ \\
$+4$ & $2{,}479$ & $0.2641_{\pm.0096}$ & $0.5676_{\pm.0185}$ \\
$+5$ & $1{,}697$ & $0.2709_{\pm.0076}$ & $0.5728_{\pm.0176}$ \\
\midrule
Overall & $19{,}486$ & $0.3623_{\pm.0075}$ & $0.6720_{\pm.0077}$ \\
\bottomrule
\end{tabular}
\end{table}

\paragraph{Aggregation and horizon-specific results}
For prefix length $n$ and future step $K$, let $N_{n,K}$ count eligible pairs and $c_{n,K}$ count hits. Horizon accuracy is
\begin{equation}
A_K = \frac{\sum_n c_{n,K}}{\sum_n N_{n,K}}.
\label{eq:rollout-aggregation}
\end{equation}
Top-1 and Top-3 use this same expression and differ only in what counts as a hit. Table~\ref{tab:rollout} reports the mean and sample standard deviation of $A_K$ over the three training trials. Overall also weights by pair count, rather than equally weighting horizons. A rally can contribute several prefixes and horizons; pairs are not independent rallies. These Type accuracies differ from the main task-averaged weighted F1.

First-step Top-1 accuracy ($0.5146\pm0.0061$) is close to the single-step result ($0.5143\pm0.0064$), but falls to $0.3142$ at $+2$ and approximately $0.26$--$0.27$ thereafter. Top-3 also falls, from $0.8501$ to $0.5456$--$0.5728$ at $+3$ to $+5$. Candidate-set membership does not establish calibrated probabilities or tactically coherent generated sequences.

Pairs decrease from $6{,}602$ at $+1$ to $1{,}697$ at $+5$, as distant targets require longer rallies. The small Top-1 rise after $+3$ is therefore not evidence of recovery on a fixed sample set. This diagnostic combines prediction feedback with changing target coverage and does not isolate their separate effects.

\subsection{Context Availability during Match Progression} \label{sec:context}
We group original table tennis test examples by available history (Table~\ref{tab:context-groups}), fixing each example to the first three observed shots and the fourth-shot target. Empty higher levels use learned fallback tokens. This is an offline stratified evaluation; examples in different groups also differ in match phase and label distribution.

\begin{table}[t]
\centering
\caption{Context-availability groups. $L_1$, the current-rally prefix, is always present; $L_2$ exists unless the rally opens its set, and $L_3$ exists unless the rally lies in the first set.}
\label{tab:context-groups}
\begin{tabular}{@{}lll@{}}
\toprule
 & First set (no $L_3$) & Later set (with $L_3$) \\
\midrule
First rally of set (no $L_2$) & Opening: $L_1$ & Set boundary: $L_1$+$L_3$ \\
Later rally (with $L_2$) & Middle: $L_1$+$L_2$ & Late: $L_1$+$L_2$+$L_3$ \\
\bottomrule
\end{tabular}
\end{table}

\begin{table}[t]
\centering
\caption{Macro F1 of hierarchy-depth variants by context availability on original table tennis (three-seed mean$\pm$std). Each example predicts the fourth shot from the first three; groups follow Table~\ref{tab:context-groups}. The full models lead once rally- and set-level history exist; the Opening column rests on $20$ samples and is not read as a ranking. Bold = column-wise best.}
\label{tab:contextmodels}
\begin{tabular}{lcccc}
\toprule
Model & Opening & Middle & Set boundary & Late \\
\midrule
Samples & $20$ & $279$ & $64$ & $989$ \\
\midrule
L1 only (TA) & $0.5217_{\pm.0354}$ & $0.5016_{\pm.0030}$ & $\mathbf{0.5160_{\pm.0111}}$ & $0.5247_{\pm.0082}$ \\
L1+L2 (TA) & $0.5373_{\pm.0225}$ & $0.5077_{\pm.0040}$ & $0.4996_{\pm.0060}$ & $0.5318_{\pm.0072}$ \\
MT-HTA Core & $\mathbf{0.5451_{\pm.0462}}$ & $0.5134_{\pm.0073}$ & $0.5019_{\pm.0226}$ & $0.5349_{\pm.0112}$ \\
Final MT-HTA & $0.5187_{\pm.0536}$ & $\mathbf{0.5143_{\pm.0066}}$ & $0.5096_{\pm.0127}$ & $\mathbf{0.5398_{\pm.0074}}$ \\
\bottomrule
\end{tabular}
\end{table}

Final has its highest mean Macro F1 in the late group ($0.5398$; Table~\ref{tab:contextmodels}), which contains $989$ of $1{,}352$ examples. The opening group contains only $20$ examples and has much greater seed variability ($0.5187\pm0.0536$). These results do not establish equal reliability across match phases. The overall majority score is not a matched baseline for individual groups, and differences between groups cannot be attributed solely to context availability.

Within the same late-group examples, mean Macro F1 is $0.5247$ for L1, $0.5318$ for L1+L2, $0.5349$ for Core, and $0.5398$ for Final (Table~\ref{tab:contextmodels}); the same ordering holds in the middle group. This pattern is consistent with benefits from available history, but the variants also differ in parameters and optional components. It is descriptive evidence, not an isolated causal estimate of higher-level context.

\subsection{Fixed-Rule Attention Case Study} \label{sec:case}

\begin{figure*}[t]
\centering
\resizebox{\textwidth}{!}{\definecolor{attn0}{HTML}{CDE2FB}
\definecolor{attn1}{HTML}{B7D3F6}
\definecolor{attn2}{HTML}{9EC5F4}
\definecolor{attn3}{HTML}{86B6EF}
\definecolor{attn4}{HTML}{6DA7EC}
\definecolor{attn5}{HTML}{5598E7}
\definecolor{attn6}{HTML}{3987E5}
\definecolor{attn7}{HTML}{2A78D6}
\begin{tikzpicture}[
    x=0.82cm,y=0.55cm,
    every node/.style={font=\footnotesize,text=black,inner sep=1pt}
]
\newcommand{\attentionpanel}[4]{%
  \begin{scope}[xshift=#1*0.82cm]
    \node[font=\small\bfseries] at (2.5,0.65) {#2};
    \foreach \cellvalue [count=\idx from 0] in {#3} {
      \pgfmathtruncatemacro{\col}{mod(\idx,5)}
      \pgfmathtruncatemacro{\row}{floor(\idx/5)}
      \pgfmathtruncatemacro{\bucket}{min(7,round(10*\cellvalue))}
      \colorlet{cellfill}{attn\bucket}
      \fill[cellfill] (\col,-\row) rectangle ++(1,-1);
      \draw[white,line width=0.4pt]
        (\col,-\row) rectangle ++(1,-1);
      \node at (\col+0.5,-\row-0.5) {\cellvalue};
    }
    \draw[black!35] (0,0) rectangle (5,-5);
    \foreach \task [count=\row from 0] in {Type,Backhand,Location,Strength,Spin} {
      \node[anchor=east] at (-0.12,-\row-0.5) {\task};
    }
    \node at (0.5,-5.5) {$L_1$};
    \ifnum#4=0
      \node at (1.5,-5.5) {$L_2^{*}$};
      \node at (2.5,-5.5) {$L_3^{*}$};
    \else
      \node at (1.5,-5.5) {$L_2$};
      \ifnum#4=1
        \node at (2.5,-5.5) {$L_3^{*}$};
      \else
        \node at (2.5,-5.5) {$L_3$};
      \fi
    \fi
    \node at (3.5,-5.5) {Hit.};
    \node at (4.5,-5.5) {Rec.};
  \end{scope}
}
\attentionpanel{0}{Opening}{0.66,0.03,0.03,0.10,0.18,0.61,0.05,0.04,0.12,0.18,0.45,0.09,0.08,0.17,0.21,0.41,0.10,0.10,0.18,0.21,0.52,0.06,0.06,0.15,0.21}{0}
\attentionpanel{7}{Middle}{0.63,0.14,0.02,0.08,0.14,0.59,0.14,0.03,0.09,0.14,0.42,0.19,0.07,0.14,0.18,0.40,0.21,0.08,0.14,0.17,0.50,0.20,0.04,0.11,0.15}{1}
\attentionpanel{14}{Late}{0.37,0.16,0.15,0.19,0.13,0.35,0.16,0.16,0.19,0.15,0.25,0.19,0.20,0.21,0.16,0.23,0.18,0.24,0.20,0.16,0.26,0.17,0.24,0.20,0.14}{2}
\node[anchor=west,font=\scriptsize] at (-1.3,-6.35)
    {$^{*}$ Learned empty summary};
\node[anchor=east,font=\scriptsize] at (11.1,-6.35) {Mean attention:};
\foreach \cellvalue [count=\idx from 0] in {0.0,0.1,0.2,0.3,0.4,0.5,0.6,0.7} {
  \pgfmathtruncatemacro{\bucket}{min(7,round(10*\cellvalue))}
  \colorlet{legendfill}{attn\bucket}
  \fill[legendfill]
      (11.3+\idx*0.84,-6.0) rectangle ++(0.84,-0.65);
  \node[font=\scriptsize] at (11.72+\idx*0.84,-6.325) {\cellvalue};
}
\end{tikzpicture}}
\caption{Final MT-HTA task-attention weights at the three fixed-rule checkpoints. Each cell is a three-seed mean, displayed to two decimals. Rows are tasks and columns are context tokens; Hit.\ and Rec.\ denote hitter and receiver. An asterisk marks an empty level, represented by a learned placeholder because no rally or set has been completed yet.}
\label{fig:context-attention}
\end{figure*}

To inspect how the context summaries are used within one match, we examine a fixed-rule case from the original table tennis test set. Among matches that contain a fourth-shot target at all three context stages, we select the one whose rally count is closest to the candidate median, breaking ties by match identifier; prediction correctness plays no part in the selection. The chosen match has 79 rallies across five sets, and the middle and late checkpoints are the first eligible rallies in search order.

All three checkpoints observe the first three shots and predict the fourth. The opening checkpoint is set 1, rally 1, with no completed rally or set. The middle checkpoint is set 1, rally 5, with four completed rallies and no completed set. The late checkpoint is set 2, rally 2, with one completed rally in the current set and one completed set. ``Late'' therefore denotes availability of all hierarchy levels, not the final portion of the match.

Figure~\ref{fig:context-attention} shows per-task attention over the hierarchy summaries and hitter/receiver context. Averaging the unrounded three-seed means across tasks, the opening checkpoint places $0.5321$ on $L_1$, compared with $0.0664$ and $0.0602$ on the empty $L_2$ and $L_3$ summaries. The nonzero empty-token weights are attention to learned fallback representations, not access to future rallies or sets. At the middle checkpoint, the mean $L_2$ weight is $0.1758$. At the late checkpoint, the weights are distributed across $L_1$ ($0.2901$), $L_2$ ($0.1708$), $L_3$ ($0.1958$), and hitter context ($0.1970$).

Prediction correctness does not increase monotonically along these checkpoints. Using three-seed mean task probabilities, Final is correct on 3 of 5 tasks at opening, 4 of 5 at middle, and 2 of 5 at late, with the true label in its top three everywhere; Core reaches 3 of 5 at the late checkpoint. These are single predictions, so the example illustrates changing context allocation without establishing a model ranking or a causal reading of attention. Table~\ref{tab:contextmodels} carries the population-level comparison.

\section{Discussion and Limitations} \label{sec:discuss}
MT-HTA provides a common architectural design for event datasets with different hierarchies, fields, and prediction tasks. The shared encoder serves the tasks within a dataset; preprocessing, configuration, and learned parameters remain dataset-specific. The evidence therefore concerns architectural applicability across racket sports. It does not measure joint training across sports, transfer to an unseen sport, or reductions in pipeline maintenance cost.

Three patterns recur across the results. Aligning the encoder to the competition hierarchy helps on every dataset, but by small amounts: Final exceeds the flat sequence baseline on all four corpora without ever pulling far ahead. The gain comes from the shared core rather than from the sport-specific refinements, since Core stays close to Final everywhere and, on original table tennis, two refinements can be removed without lowering mean Macro F1. Classification and spatial accuracy respond differently to the same components, so an aggregate score hides which outputs a change helps. The hierarchy is therefore the part that ports across sports, while the refinements are what a new source should expect to re-tune.
 
The accuracy margins are modest and estimated from three training seeds on one fixed split per dataset. Match-level splitting does not test generalization to entirely unseen players. Differences in task sets, label distributions, hierarchy depth, and dataset composition also prevent attributing cross-sport score differences to structural richness. The detailed component analysis is limited to original table tennis.

Location remains difficult, and classification and distance metrics can favor different configurations. Location distance is averaged only over examples whose target and prediction both lie in the grid, so each model is scored on a different subset. Lower distance alone does not establish better location prediction. Grid discretization and imbalanced placement labels are additional limitations; continuous-coordinate modeling could reduce discretization error, but would require suitable annotations and separate evaluation.
 
The original ShuttleNet cannot be applied directly here: it assumes a fixed prefix length and targets stroke-sequence generation, whereas our setting has dynamic history length and single-step multi-task prediction. The Adapted ShuttleNet, despite keeping most components identical to the original design, still differs from its original stroke-generation setting in output representation and decoding, and uses only the current-rally prefix. Its scores may reflect adaptation and context differences.

\section{Conclusion} \label{sec:conclusion}
We presented MT-HTA, a hierarchy-aligned multi-task Transformer for next-shot prediction over heterogeneous racket-sport event data. The same architectural design supports three- and four-level match structures and different task sets, with separate training on each dataset. Across four datasets, Final MT-HTA achieves competitive task-averaged F1 and the highest means among the compared main models, with modest margins and an effective tie on badminton. Ablations on original table tennis support hierarchical encoding with task attention and complementary feature-token interaction, while exposing trade-offs between classification and spatial error.
 
Future work includes continuous-coordinate location prediction, objective physical measurements for attributes such as strength, multimodal pre-shot cues, and measured transfer between sports. Evaluating incremental processing and live annotation latency would further establish the framework's suitability for in-match analysis.

\section*{Acknowledgments and GenAI Usage Disclosure}
This work was supported by the National Science and Technology Council of Taiwan under Grant 113-2221-E-008-100-MY3.
Computational and storage resources were provided by the National Center for High-performance Computing of National Applied Research Laboratories, Taiwan.
The authors largely used LLMs to improve language and readability. The authors reviewed and edited the content as needed. The authors take full responsibility for the content.

\bibliographystyle{apalike}  
\bibliography{ref}  

\end{document}